\documentclass[11pt]{article}

\usepackage[preprint]{acl}

\usepackage{times}
\usepackage{latexsym}

\usepackage[T1]{fontenc}

\usepackage[utf8]{inputenc}

\usepackage{microtype}

\usepackage{inconsolata}

\usepackage{graphicx}

\usepackage{amsfonts}       
\usepackage{amsmath}
\usepackage{subcaption}
\usepackage[table]{xcolor}            
\definecolor{lightred}{HTML}{FFCCCC}  
\definecolor{lightblue}{HTML}{CCE5FF} 
\usepackage{wrapfig}    
\usepackage{booktabs}       

\usepackage{multirow}
\usepackage{tcolorbox}
\newtcolorbox{takeaway}{
  colback=blue!5,
  colframe=blue!40,
  boxrule=0.5pt,
  arc=2pt,
  left=6pt,
  right=6pt,
  top=6pt,
  bottom=6pt
}

\title{Charting Empirical Laws for LLM Fine-Tuning in Scientific Multi-Discipline Learning}

\author{
 \textbf{Lintao Wang\textsuperscript{1,2,\dag}},
 \textbf{Zhuqiang Lu\textsuperscript{1,\dag}},
 \textbf{Yilin Zhu\textsuperscript{1,\dag}},
 \textbf{Kun Hu\textsuperscript{3}},
 \\
 \textbf{Zhenfei Yin\textsuperscript{4}},
 \textbf{Shixiang Tang\textsuperscript{2,5}},
 \textbf{Zhiyong Wang\textsuperscript{1}},
 \textbf{Wanli Ouyang\textsuperscript{2,5}},
 \textbf{Xinzhu Ma\textsuperscript{2,6,*}}
 \\
 \textsuperscript{1}The University of Sydney,
 \textsuperscript{2}Shanghai AI Laboratory,
 \textsuperscript{3}Edith Cowan University,
 \\
 \textsuperscript{4}University of Oxford,
 \textsuperscript{5}The Chinese University of Hong Kong,
 \textsuperscript{6}Beihang University
 \\
   {\small\textsuperscript{\dag}Equal contribution. \quad \textsuperscript{*}Corresponding authors.}
 }

\begin{document}
\maketitle
\begin{abstract}
While large language models (LLMs) have achieved strong performance through fine-tuning within individual scientific domains, their learning dynamics in \textit{multi-disciplinary} contexts remains poorly understood, despite the promise of improved generalization and broader applicability through cross-domain knowledge synergy. 
In this work, we present the first systematic study of multi-disciplinary LLM fine-tuning, constructing a five-discipline corpus and analyzing learning patterns of full fine-tuning, LoRA, LoRA-MoE, and LoRA compositions. 
Particularly, our study shows that multi-disciplinary learning is substantially more variable than single-discipline training and distills four consistent empirical laws: (1) \textbf{Balance-then-Diversity}: low-resource disciplines degrade performance unless mitigated via diversity-aware upsampling; (2) \textbf{Merge-then-Align}: restoring instruction-following ability is critical for cross-discipline synergy; (3) \textbf{Optimize-then-Scale}: parameter scaling offers limited gains without prior design optimization; and (4) \textbf{Share-then-Specialize}: asymmetric LoRA-MoE yields robust gains with minimal trainable parameters via shared low-rank projection.
Together, these laws form a practical recipe for principled multi-discipline fine-tuning and provide actionable guidance for developing generalizable scientific LLMs.
\end{abstract}

\section{Introduction}

Artificial intelligence (AI) has made significant strides in scientific discovery, supporting tasks such as hypothesis generation, experimental design, and data analysis across disciplines \citep{xie2024sciliterature, ai4science2023impact, wang2023scidis, zhang2024sciLLM}.  
Prior AI4Science works built specialized models to solve specific scientific problems \citep{lamurias2023metagenomic,pei2024fabind,laghuvarapu2024codrug,kohler2023rigid,hao2023gnot}. While these models show promising performance on specific tasks, they are often constrained by custom architectures and small, specialized labeled datasets, which restrict their generalization capabilities.  
To address these limitations, large-scale pre-training has emerged as a promising approach for building scientific foundation models \citep{ji2021dnabert, chen2023fengwu, chen2024uni}. The scale of pre-training data enables these models to surpass specialized counterparts on downstream tasks. 
More recently, inspired by the remarkable capabilities of large language models (LLMs) \citep{zhao2023llmsurvey}, studies have adapted general-purpose LLMs to scientific tasks by fine-tuning them with scientific instructions \citep{yue2023mammoth,fang2023mol,han2023medalpaca, zhang2024chemllm}, effectively transferring problem-solving abilities to scientific problems and achieving impressive results. 

Despite recent progress in AI4Science, developing models with multi-disciplinary knowledge remains underexplored. Such models can capture cross-domain synergies, support broader generalization, and enable complex scientific reasoning beyond the scope of single-discipline approaches. Recent efforts have begun to address this gap. 
UniSTD \citep{tang2025unistd} develops a spatio-temporal foundation model to enhance performance across four spatio-temporal disciplines. 
SciReasoner \citep{wang2025scireasoner} unifies representations of proteins, nucleotides, molecules and materials via natural language, enabling reasoning-driven modeling for biology and chemistry. 
Instead of training from scratch, X-LoRA \citep{buehler2024xlora} combines low-rank adaptation adapters (LoRA) fine-tuned on individual disciplines through a mixture-of-experts framework, leveraging single-discipline knowledge for multi-disciplinary tasks.
However, these approaches adopt disparate algorithmic design choices without systematically analyzing the differences between multi- and single-discipline learning, limiting their generalizability and making it challenging to extend their designs to other domains.

In this study, we systematically investigate learning dynamics across single- and multi-discipline settings, providing insights into how to develop more generalizable scientific LLMs. 
We curate a text-based multi-discipline corpus spanning five scientific domains: mathematics, chemistry, biology, medicine, and geography. 
Leveraging this corpus, we investigate four representative fine-tuning strategies: full fine-tuning \citep{zhou2023lima}, LoRA \citep{hu2022lora}, LoRA-MoE \citep{zadouri2023molora}, and LoRA composition \citep{buehler2024xlora}, applied to Qwen2.5 7B Instruct \citep{yang2024qwen2.5} under varying data scales. 
Evaluation on in-domain benchmarks reveals that multi-discipline learning trajectories diverge substantially from those in single-discipline settings. In particular, disciplines with limited data experience greater instability, cross-discipline synergy is inconsistent, and multi-discipline fine-tuning can degrade discipline-specific performance in terms of average accuracy.  
We further analyze these phenomena to uncover the empirical regularities that govern multi-discipline fine-tuning and distill them into four laws that together form a practical recipe for building robust scientific LLMs:
\begin{itemize}
\item \textbf{Balance-then-Diversity:} Low-resource disciplines disproportionately hinder multi-discipline learning, leading to reduced accuracy and increased variance. We show that diversity-aware up-sampling, rather than naïve duplication, mitigates this issue by maintaining both balance and diversity in cross-discipline data contributions.
\item \textbf{Merge-then-Align:} Instruction-following ability is a prerequisite for effective cross-discipline transfer. Multi-discipline fine-tuning alone degrades alignment, while mixing a modest amount of general instruction data restores instruction-following and unlocks synergy across disciplines.
\item \textbf{Optimize-then-Scale:} Simply enlarging the number of trainable parameters provides negligible gains in multi-discipline settings. Performance improvements instead depend on careful model design and optimization choices, suggesting that scaling should follow—not precede—architectural optimization.
\item \textbf{Share-then-Specialize:} LoRA-MoE architectures with asymmetric parameter sharing (e.g., shared A matrices) first promote effective cross-discipline knowledge sharing, and then enable stable expert specialization, yielding robust multi-discipline improvements comparable to full fine-tuning while using only a small fraction of the parameters.
\end{itemize}

\section{Related Work}
\textbf{AI4Science} applies artificial intelligence to advance scientific research through both specialized task models \citep{lamurias2023metagenomic,laghuvarapu2024codrug,hao2023gnot} and large-scale discipline-wise scientific foundation models \citep{ji2021dnabert,chen2023fengwu,chen2024uni,price2025gencast,zeni2025mattergen}. 
Large language models (LLMs) have also emerged as scientific foundation models, leveraging their advanced reasoning and language understanding capabilities to support scientific problem solving. Various disciplines have developed their own domain-specific LLMs, including models for mathematics \citep{cobbe2021gsm8k, toshniwal2024openmathinstruct2}, biology \citep{fang2023mol}, medicine \citep{li2023chatdoctor, li2024llava-med}, chemistry \citep{zhang2024chemllm, li20243dmolm}, and geoscience \citep{deng2024k2, ma2024weatherqa}. However, these models were typically trained and evaluated within isolated domains, neglecting the inherently interdisciplinary nature of many scientific problems. Recent studies \citep{tang2025unistd, xia2025naturelm, wang2025scireasoner,buehler2024xlora} attempted multi-disciplinary modeling, yet these approaches adopt diverse strategies and lack systematic investigation into the learning dynamics and challenges inherent to multi-disciplinary fine-tuning.
In this work, we aim to bridge this gap by providing a comprehensive analysis of multi-disciplinary fine-tuning paradigms. 

\begin{figure*}[t]
    \centering
    \includegraphics[width=\linewidth]{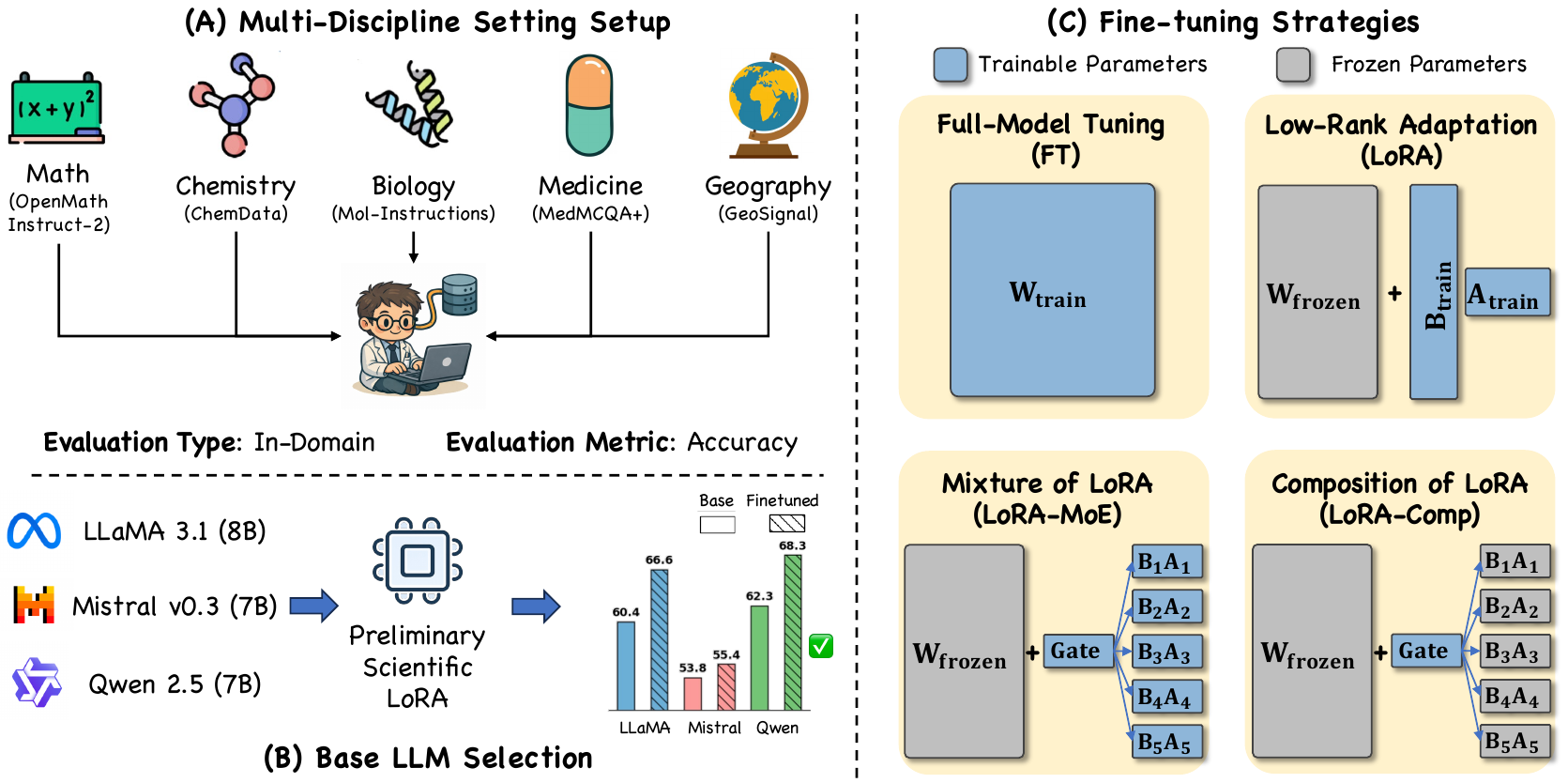}
    \caption{\textbf{Overview of the Empirical Study Methodology}. (A) Five scientific disciplines included in our multi-discipline setting. (B) Selection of Qwen 2.5 based on superior pre-na dn post-tuning scientific performance. (C) Fine-tuning strategies investigated by distinguishing trainable and frozen parameters.}
    \label{fig:overview}
\end{figure*}

\noindent
\textbf{LLM Fine-Tuning} adapts large language models to specific tasks, with instruction tuning \citep{wei2021zero-shotlearner, ouyang2022instructgpt, zhang2023instruction} widely used to align models via curated instruction-output pairs. 
As model sizes grow, the high cost of full fine-tuning has driven the development of parameter-efficient fine-tuning (PEFT) methods that update only a small subset of parameters.
These methods introduces lightweight trainable modules \citep{houlsby2019parameter, liu2022ia3, lester2021promttuning, li2021prefix} or selectively update existing parameters \citep{zaken2021bitfit}. A prominent PEFT approach is low-rank adaptation (LoRA) \citep{hu2022lora} and its variants \citep{zhang2023adalora, liu2024dora}, which reduce parameter cost by factorizing weight updates. 
In this work, we analyze both full fine-tuning and PEFT in the context of scientific LLMs.

\noindent
\textbf{Multi-Task Learning} (MTL) improves performance by enabling knowledge sharing across tasks, closely aligning with multi-discipline learning. 
Traditional MTL shares parameters \citep{liu2019end, zhao2024layer, ruder2019latent} or balances task gradients \citep{chen2018gradnorm, liu2023famo, senushkin2023independent}. 
In the LLM context, MTL has been adapted using parameter-efficient fine-tuning (PEFT).  
One line of research composes task-specific LoRA modules \citep{zhang2023composing, ostapenko2024towards, prabhakar2024lorasoups, buehler2024xlora}, using arithmetic operations \citep{zhang2023composing} or dynamic gating \citep{buehler2024xlora}.
Another line integrates LoRA with mixture-of-experts (MoE) architectures, where LoRA-MoE \citep{zadouri2023molora, luo2024moelora} achieves competitive performance. Extensions include asymmetric sharing in HydraLoRA \citep{tian2024hydralora} and hierarchical routing in HMoRA \citep{liao2025hmora}. 
We focus on two representative PEFT-based MTL paradigms: LoRA composition and LoRA-MoE.

\begin{table*}[t]
  \centering
  \small
  \setlength{\tabcolsep}{6pt}
  \begin{tabular}{p{2.2cm}<{\raggedright} r r c c c}
    \toprule
    \multirow{2}{*}{\textbf{Discipline}} &
    \multicolumn{2}{c}{\textbf{Data Scale}} &
    \multirow{2}{*}{\textbf{Source (Train / Eval)}} &
    \multicolumn{2}{c}{\textbf{Text Statistics}} \\
    \cmidrule(lr){2-3}
    \cmidrule(lr){5-6}
    & Samples & \% &
    & Avg. Length & Unique Tokens \\
    \midrule
    Math        & 2,000,000 & 60.7 & \citet{toshniwal2024openmathinstruct2} / \citet{cobbe2021gsm8k} & 267.74 & 1,995,829 \\
    Chemistry  &   713,218 & 21.6 & \citet{zhang2024chemllm} / \citet{zhang2024chemllm}               &  57.66 & 1,232,633 \\
    Biology    &    51,427 &  1.6 & \citet{fang2023mol} / \citet{fang2023mol}                       &  50.61 &    22,574 \\
    Medicine   &   490,766 & 14.9 & \citet{pal2022medmcqa}, etc. / \citet{pal2022medmcqa}                 & 139.17 &   529,564 \\
    Geography  &    39,749 &  1.2 & \citet{deng2024k2} / \citet{deng2024k2}                         & 114.20 &    60,209 \\
    \midrule
    \textbf{Total} & \textbf{3,295,160} & \textbf{100.0} & -- & \textbf{188.17} & \textbf{3,840,809} \\
    \bottomrule
  \end{tabular}
  \caption{
  \textbf{Multi-discipline corpus statistics.}
  Average length is measured in words per sample and
  unique tokens are defined as tokens with fewer than 10 co-occurrences across disciplines.
  }
  \label{tab:data-stats}
\end{table*}

\section{Experiment Setup} 
\label{sec:setup}

\textbf{Multi-Discipline Setting.}
To emulate the diversity in quantity and format characteristic of real-world heterogeneous multi-disciplinary data, we construct a multi-discipline training dataset and evaluation benchmarks by aggregating existing open-source datasets from individual disciplines (Figure \ref{fig:overview} and Table \ref{tab:data-stats}). Specifically, our study encompasses five disciplines: math, chemistry, biology, medicine, and geography.
For mathematics, we adopt a 2M-sample subset from OpenMathInstruct-2 \cite{toshniwal2024openmathinstruct2}, which is augmented upon MATH \cite{hendrycks2021mathdataset} and GSM8K \cite{cobbe2021gsm8k} to support large-scale supervised fine-tuning for mathematical reasoning. 
For chemistry, we use ChemData \cite{zhang2024chemllm} containing 700K instruction-based QA pairs on molecular properties, reactions, and related tasks. 
For biology, we select the biomolecular text subset of Mol-Instructions \cite{fang2023mol} for training. 
For medicine, we compile a large-scale dataset by aggregating samples from MedMCQA \cite{pal2022medmcqa}, MedAlpaca \cite{han2023medalpaca}, ChatDoctor \cite{li2023chatdoctor}, MedInstruct-52K \cite{zhang2023alpacare}, and others (Appendix \ref{sec:add}). 
For geography, we adopt GeoSignal \cite{deng2024k2}, which includes QA covering geoscientific topics. 

\noindent
\textbf{Evaluation Setting.} 
To assess discipline-specific knowledge acquired through scientific fine-tuning, we adopt in-domain evaluation benchmarks tailored to each discipline. Specifically, mathematics is evaluated using GSM8K test set \citep{cobbe2021gsm8k}, while chemistry is assessed with ChemBench \citep{zhang2024chemllm}, covering nine representative chemical tasks. For biology, we employ the biomolecular multiple-choice subset of Mol-Instruction test set \citep{fang2023mol}. Medicine is evaluated using MedMCQA test set \cite{pal2022medmcqa}, and geography is assessed using the multiple-choice subset of GeoBench \cite{deng2024k2} to ensure consistent evaluation formats.
We conduct all evaluations using the lm-evaluation-harness framework \citep{eval-harness}, employing accuracy as the primary metric.

\noindent
\textbf{Fine-Tuning Setting.} We study the learning patterns for the full fine-tuning and three parameter efficient tuning methods (PEFT) (Figure \ref{fig:overview}):

\begin{itemize}
    \item \textbf{Full-Model Tuning (FT)}: This standard approach updates all parameters of the language model. Specifically, each forward operation $\mathbf{x}_{\text{out}} = \mathbf{W}\mathbf{x}_{\text{in}}$ will have the pre-trained weight matrix $\mathbf{W}$ fully trainable. 
    \item \textbf{Low-Rank Adaptation (LoRA)}: LoRA introduces trainable low-rank matrices $\mathbf{B} \in \mathbb{R}^{r \times d_{\text{out}}}$ and $\mathbf{A} \in \mathbb{R}^{d_{\text{in}} \times r}$ to adapt pre-trained weights $\mathbf{W} \in \mathbb{R}^{d_{\text{in}} \times d_{\text{out}}}$, where $r \ll \min(d_{\text{in}}, d_{\text{out}})$. During fine-tuning, the model uses $\mathbf{W} + \mathbf{BA}$, updating only $\mathbf{B}$ and $\mathbf{A}$.
    \item \textbf{Mixture of Low-Rank Adaptation (LoRA-MoE):} LoRA-MoE extends LoRA by introducing a set of trainable low-rank matrices $\{\mathbf{B}_i\}_{i=0}^k$ and $\{\mathbf{A}_i\}_{i=1}^k$, where $k$ denotes the number of experts. The adapted weight is computed as $\mathbf{W} + \sum_{i=1}^{k} \omega_i \cdot \mathbf{B}_i\mathbf{A}_i$, with gating weights $\omega_i$ dynamically predicted by a gating network. The gating network is typically implemented as a lightweight MLP, $\omega = f_{\text{MLP}}(\mathbf{x}_{\text{in}})$, and is jointly optimized with the low-rank matrices.
    \item \textbf{Composition of Low-Rank Adaptation (LoRA-Comp):} LoRA-Comp adopts a similar formulation to LoRA-MoE, where the adapted weight is composed of multiple expert LoRA modules. Differently, the expert LoRA modules are pre-trained on individual discipline-specific corpora, and during multi-discipline fine-tuning, only the gating network is optimized to dynamically combine the pre-trained single-discipline LoRA experts. 
\end{itemize}

For full-model tuning, we fine-tune all model parameters using a learning rate of $7 \times 10^{-6}$, weight decay of $0.1$, and a warm-up ratio of $0.05$ for 1 epoch. 
For PEFT methods, we adopt a learning rate of $1 \times 10^{-4}$, weight decay of $0.01$, and a warm-up ratio of $0.1$, with training conducted for 1 epoch. We set LoRA rank as 16 and number of experts for LoRA-MoE and LoRA-Comp match the number of 5 disciplines. See Appendix \ref{sec:add} for details.

\noindent
\textbf{Base LLMs.} 
We use three 7B-scale instruction-tuned LLMs—LLaMA 3.1 \citep{grattafiori2024llama3}, Mistral v0.3 \citep{mistral0.3}, and Qwen2.5 \citep{yang2024qwen2.5} as base models for preliminary scientific fine-tuning using LoRA \citep{hu2022lora} under identical settings.
As shown in Figure \ref{fig:overview} (B), scientific fine-tuning performance is highly sensitive to the choice of the pre-trained base model and does not consistently yield gains across tasks. 
Among the evaluated models, Qwen2.5 7B Instruct achieves the strongest pre-trained performance and the largest improvements after fine-tuning. We therefore adopt Qwen2.5 7B Instruct for all subsequent experiments.
More evaluation results can be found in Appendix \ref{sec:base}.

\section{Empirical Laws for Multi-Discipline LLM Fine-Tuning}

\begin{figure*}[t]
    \centering
    \includegraphics[width=\linewidth]{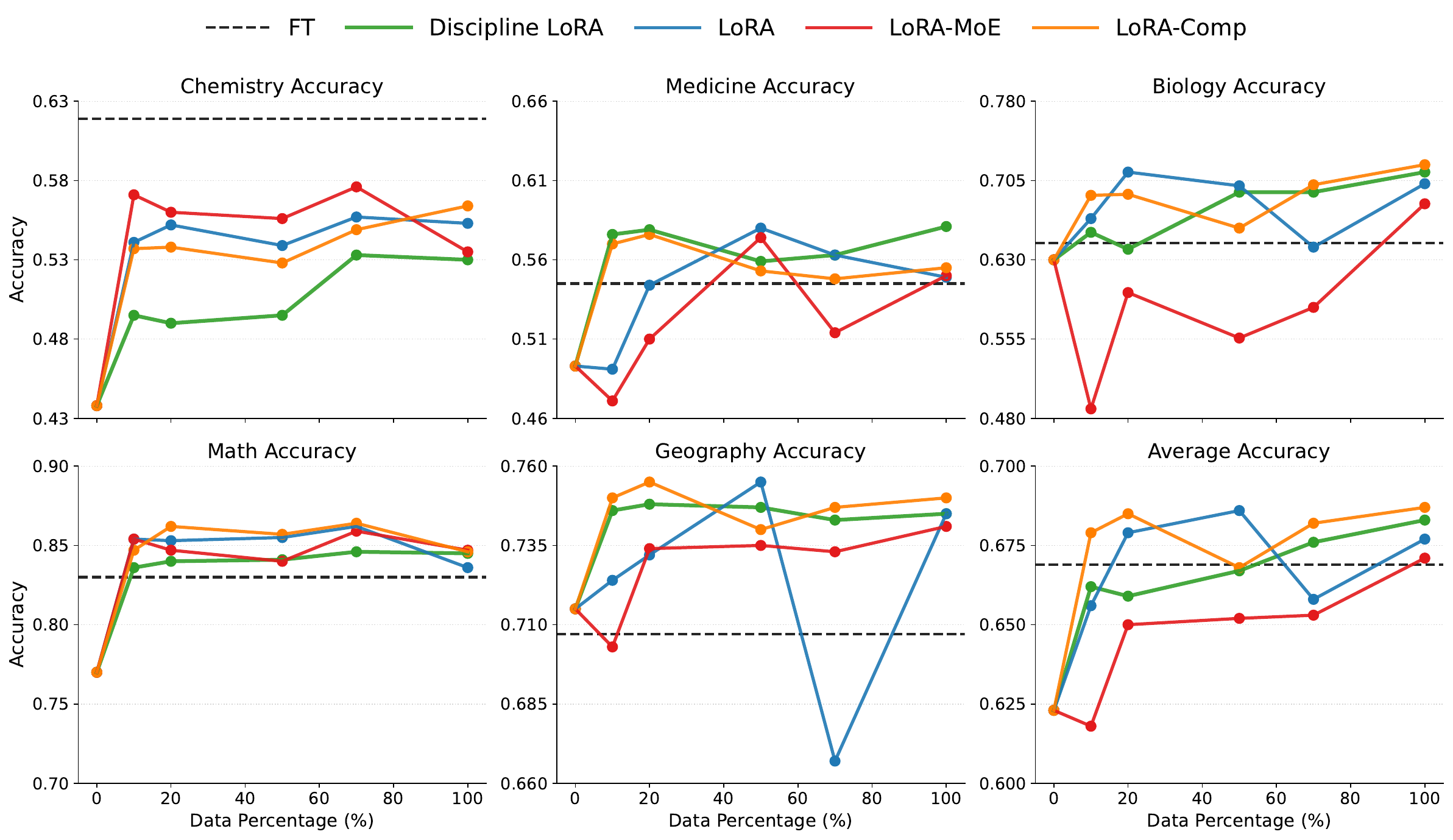}
    \caption{\textbf{Learning Dynamics of Fine-tuning Strategies.} Multi-disciplinary fine-tuning is less effective and stable than single-discipline tuning, highlighting trade-offs between generality, performance, and stability across methods.}
    \label{fig:pattern}
\end{figure*}

To compare single- and multi-discipline learning, we analyze their learning patterns across fine-tuning strategies and data scales (Figure~\ref{fig:pattern}). In the single-discipline setting, performance scales predictably with data size where modest amounts already yield strong results with continued improvements as data increases \citep{zhou2023lima, zhang2024scalingft}. In contrast, multi-discipline learning exhibits markedly less stable scaling behavior.

\noindent
\textbf{Fine-Tuning Perspective.} Scaling curves exhibit greater variance as data volume changes, a pattern that is particularly pronounced in data-scarce fields such as medicine, biology, and geography.  While certain domains (e.g., chemistry) sometimes benefit from cross-disciplinary data, these gains are inconsistent. On average, multi-disciplinary fine-tuning tends to degrade discipline-specific performance, as reflected in lower average accuracy compared to single-discipline fine-tuning. Furthermore, directly applying full-model fine-tuning (FT) on multi-discipline data results in sub-optimal outcomes. In this setting, conflicting learning signals across disciplines and overfitting to dominant data sources could occur, which not only undermines generalization but also leads to performance inferior to PEFT methods, despite FT involving substantially more trainable parameters.

\noindent
\textbf{Architectural Perspective.} Fine-tuning strategies also differ in stability. LoRA-Comp, training a lightweight router on top of discipline-specific LoRA adapters, yields the most stable trajectories and most closely resembles single-discipline behavior. Nonetheless, the router’s limited capacity constrains its ability to model cross-disciplinary interactions, producing lower peak performance than LoRA and LoRA-MoE models trained from scratch on aggregated data.  Because they are initialized from scratch in the multi-disciplinary regime, LoRA and especially LoRA-MoE exhibit greater performance variability and generally underperform their single-discipline counterparts.

\begin{figure*}[t]
    \centering
    \includegraphics[width=0.95\linewidth]{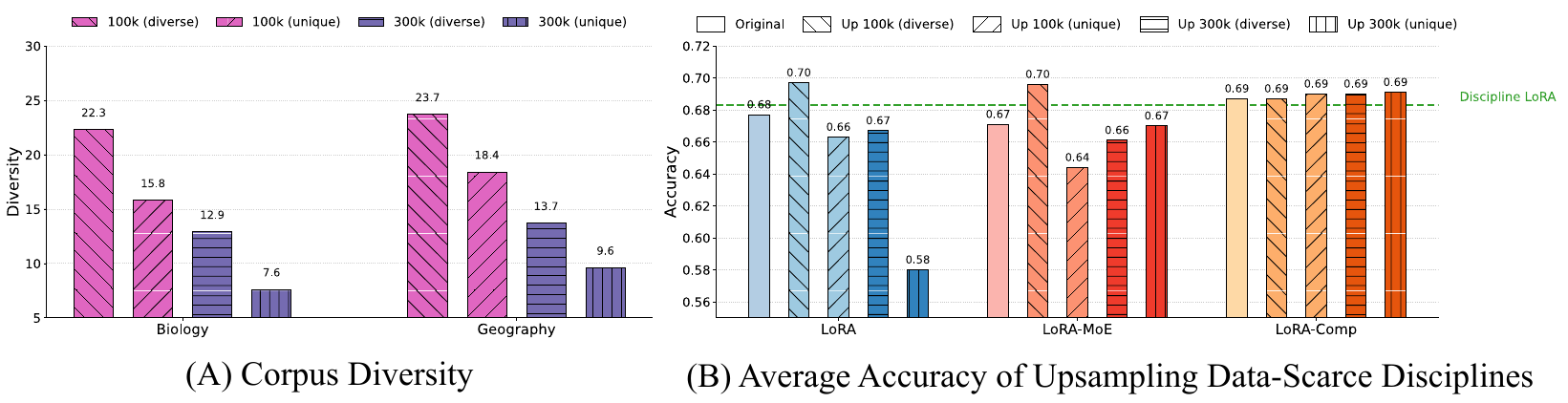}
    \caption{Analysis of how data-scarce discipline affects the multi-disciplinary learning. }
    \label{fig:upsample and diversity}
\end{figure*}

\subsection{Balance-then-Diversity Empirical Law}

\begin{takeaway}
\textbf{Balance-then-Diversity.}
Multi-discipline learning first requires balancing data-scarce disciplines to reduce instability, and further gains depend on preserving sample diversity rather than simply increasing data volume.
\end{takeaway}

We begin by examining the impact of data-scarce disciplines on multi-discipline learning.
As shown in Table~\ref{tab:data-stats}, dataset sizes vary significantly across disciplines given its real-world heterogeneity.
Figure~\ref{fig:pattern} reveals that geography and biology have high variance in performance, in contrast to the stable trends observed in chemistry and mathematics.
We hypothesize that this instability arises from difficulty in learning fine-grained knowledge of low-resource domains. 
Given the cost of acquiring additional scientific data, we evaluate two straightforward upsampling methods to expand scarce data: diverse upsampling, which randomly duplicates samples, and unique upsampling, which selectively duplicates samples containing discipline-specific tokens. 
These strategies are applied to biology and geography, scaled to 100K and 300K samples, while other disciplines remain unchanged. 

As shown in Figure~\ref{fig:upsample and diversity}(B), upsampling data-scarce disciplines improves overall performance, indicating that low-resource disciplines hinder multi-discipline learning. However, gains do not scale consistently with data volume, and diversity-aware upsampling consistently outperforms unique-token-based strategies.
This effect stems from reduced sample diversity under large-scale and unique-token-based upsampling, which induces overfitting to repeated prompts. Consistently, the unique n-gram diversity metric~\citep{song2024diversity} (Figure~\ref{fig:upsample and diversity}(A)) shows that diversity degradation strongly correlates with performance drops. LoRA-Comp remains relatively stable, likely due to its limited trainable parameters constraining overfitting.
Overall, upsampling low-resource disciplines benefits both upsampled and non-upsampled domains (see Appendix \ref{sec:add_res}), highlighting a balance-then-diversity principle: effective multi-discipline learning requires first correcting data imbalance and then preserving sample diversity.

\begin{figure*}[t]
    \centering
    \includegraphics[width=\linewidth]{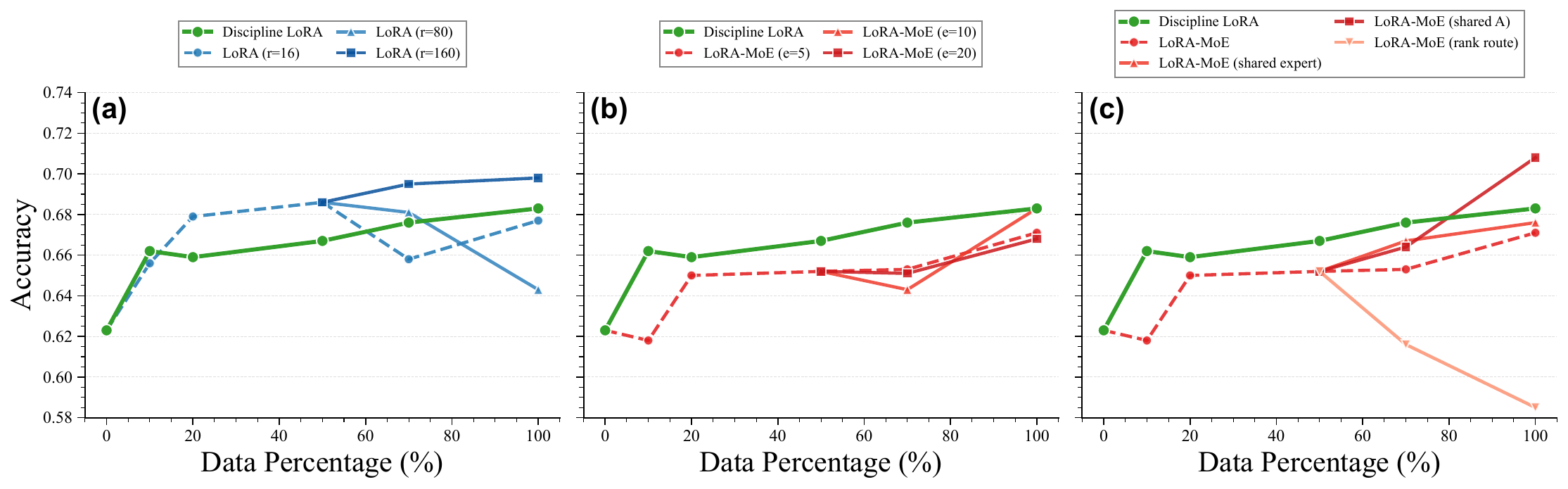}
    \caption{(a, b) Scaling LoRA rank and LoRA-MoE experts; (c) Different MoE designs.}
    \label{fig:analysis2}
\end{figure*}

\subsection{Merge-then-Align Empirical Law}

\begin{takeaway}
\textbf{Merge-then-Align.}
Multi-discipline training should first merge heterogeneous scientific data and then realign the model with general instruction-following data to restore alignment and improve overall multi-discipline synergy.
\end{takeaway}

\begin{table}[t]                  
  \centering
  \resizebox{0.99\linewidth}{!}{
  \begin{tabular}{l|p{1.7cm}<{\centering}|cc}
  \toprule
  \textbf{Methods} & \textbf{IF} (\%)& \textbf{Science} & \textbf{IFEVal} \\
  \midrule
      Qwen 2.5 & - & 0.623 & 0.808\\
  \midrule
    Dis. LoRA & - & 0.683 & - \\
  \midrule
  \multirow{3}{*}{LoRA}&0&0.677&0.725\\
  &70& 0.708 &0.742\\
  &100&\textbf{0.712}&\textbf{0.748}\\
  \midrule
  \multirow{3}{*}{LoRA-MoE}&0&0.671&0.763\\
  &70& \textbf{0.712} &0.730\\
  &100&0.708&\textbf{0.764}\\
  \midrule
  \multirow{3}{*}{LoRA-Comp}&0&0.687&0.769\\
  &70& \textbf{0.726 }&\textbf{0.783}\\
  &100&0.692&0.773\\
  \bottomrule
  \end{tabular}}
  \caption{Model performance mixing additional IF data.}
  \label{tab:instructfollow}
\end{table}

Multi-discipline fine-tuning requires adapting pre-trained LLMs to heterogeneous scientific corpora, which can degrade their instruction-following capability for effective task generalization. We hypothesize that this degradation contributes to the suboptimal learning patterns observed in multi-discipline settings.
To evaluate this, we assess models on the IFEval benchmark \citep{zhou2023instruction}, which measures instruction-following ability. As shown in Table~\ref{tab:instructfollow}, all fine-tuned models underperform relative to the base LLM, confirming a decline in alignment quality.

To address this issue, we augment the multi-discipline training corpus with general-domain instruction-following data from InternLM \citep{cai2024internlm2}, and fine-tune new models on this mixed dataset with varying percentage. This consistently improves instruction-following performance and, importantly, yields stronger results on multidisciplinary evaluation benchmarks.
Overall, these results support a merge-then-align principle: first merging diverse domain data to acquire broad knowledge, followed by explicit alignment to restore and strengthen instruction-following capability for effective multi-discipline generalization.

\subsection{Optimize-then-Scale Empirical Law}

\begin{takeaway}
\textbf{Optimize-then-Scale.}
The model design choices should be prioritized, as scaling simply yields limited or even negative returns.
\end{takeaway}

Compared to single-discipline fine-tuning, the multi-discipline setting requires the model to process and integrate a broader and more diverse set of data, potentially increasing optimization difficulty. We hypothesize that part of the observed instability in multi-discipline fine-tuning may stem from insufficient trainable parameter capacity. 
To examine this hypothesis, we systematically scale the trainable parameters of LoRA and LoRA-MoE by increasing their ranks and numbers of experts, respectively.
However, as shown in Figure \ref{fig:analysis2} (a,b), enlarging the trainable parameter size yields only marginal improvements in multi-discipline performance and, in some cases, leads to inverse scaling effects (e.g., LoRA with rank 80). 
These results are consistent with prior observations~\citep{zhang2024scalingft} that scaling PEFT capacity alone does not guarantee improved outcomes.
Overall, these results indicate that multi-discipline fine-tuning is constrained less by parameter capacity than by optimization and architectural factors, underscoring an optimize-then-scale principle: robust optimization strategies must be established before increasing model or adaptation scale can be effective.

\begin{figure*}[t]
    \centering
    \includegraphics[width=\linewidth]{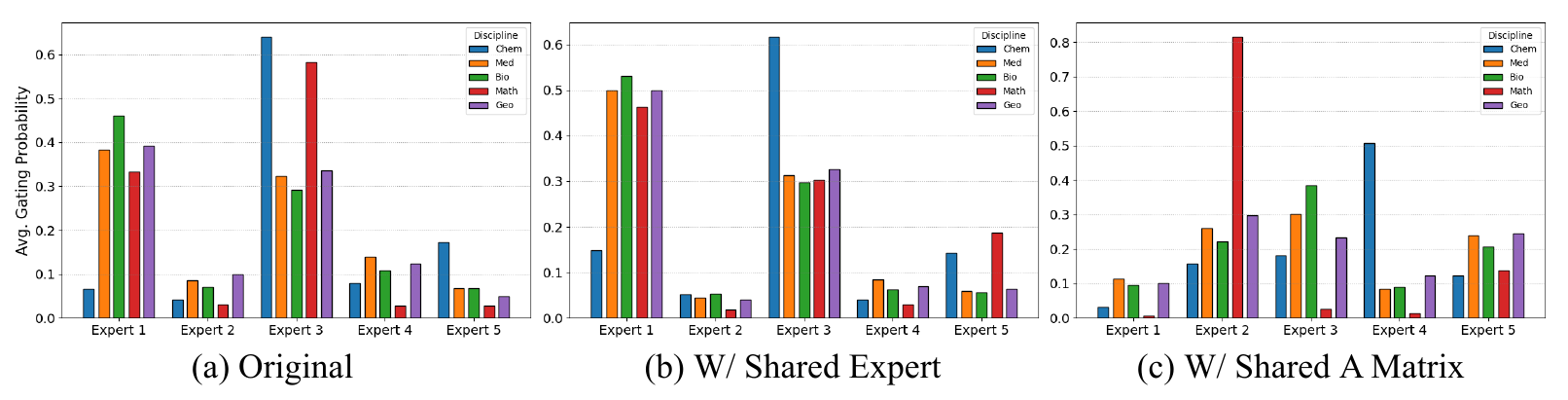}
    \caption{Mean expert routing probabilities computed from intermediate activations at the final feedforward layer.}
    \label{fig:expert}
\end{figure*}

\begin{table*}[t]
  \centering
  \resizebox{1\linewidth}{!}{
  \begin{tabular}{l|ccccccc c c c cc}
    \toprule
    \multirow{2}{*}{\textbf{Method}}&\multicolumn{7}{c}{\textbf{In-Domain}}&&\textbf{Out-Domain}&&\multirow{2}{*}{\textbf{Param.} (\%) }&\multirow{2}{*}{\textbf{GPU Hours}}
    \\
    \cmidrule{2-8}  \cmidrule{10-10}
      & Chem
      & Med 
      & Bio
      & Math
      & Geo
      & Avg
      & $\Delta m (\%)$
      && MMLU \\
    \midrule
    QWen 2.5 7B Instruct& 0.509 & 0.493 & 0.630 & 0.770 & 0.715 & 0.623 & - && 0.688 && 0 & - \\
    \midrule
    Discipline LoRA
      & 0.530 & 0.581 & 0.713 & \underline{0.845} & \underline{0.745} & 0.683 & 0 && - && 0.264 $\times$ 5 & - \\
    \midrule
    FT (ori.)
      & \underline{0.619} & 0.545 & 0.646 & 0.830 & 0.707  &  0.669 & \cellcolor{lightblue} -1.135 && 0.719 && 100 & 1920 \\
    FT (tuned)
      & \textbf{0.649} & \underline{0.588} & \underline{0.821} & 0.831 & 0.743  &  \underline{0.726} & \cellcolor{lightred} +7.376 && 0.707 && 100 & 2080 \\
    LoRA-MoE (ori.)
      & 0.535 & 0.550 & 0.683 & \textbf{0.847} & 0.741  & 0.671 & \cellcolor{lightblue} -1.780 && \underline{0.732} && 1.315 & 344 \\
    LoRA-MoE (tuned)
      & 0.613 & \textbf{0.603} & \textbf{0.823} & 0.839 & \textbf{0.760}  & \textbf{0.728} & \cellcolor{lightred} +7.236 && \textbf{0.739} && 0.902 & 328 \\
    \bottomrule
  \end{tabular}
  }
  \caption{Results comparison of our recipe-tuned tuned multi-discipline fine-tuning methods.}
  \label{tab:recipe}
\end{table*}

\subsection{Share-then-Specialize Empirical Law}

\begin{takeaway}
\textbf{Share-then-Specialize.}
Multi-discipline performance improves robustly when knowledge sharing is enforced latently through asymmetric architectures before enabling selective knowledge specialization.
\end{takeaway}

Despite the success of LoRA-MoE in general MTL \citep{tian2024hydralora, lin2024teamlora, chen2023octavius}, it surprisingly underperforms in the multi-discipline setting (Figure \ref{fig:pattern}). Given the numerous LoRA-MoE variants proposed to enhance performance, we hypothesize that improved architectural design could better support multi-discipline learning. We therefore investigate two key components: expert design and gating network design. 
For expert design, inspired by DeepSeek-MoE \citep{liu2024deepseekv3}, we introduce an additional shared LoRA expert alongside the five discipline-specific experts to promote shared and specialized knowledge separation. Additionally, following HydraLoRA \citep{tian2024hydralora}, we evaluate an asymmetric design that shares the LoRA A matrix while routing only the B matrices, encouraging cross-discipline commonality. 
For the gating network, we specifically explore rank-wise routing \citep{ning2024mode}, where individual routing weights are assigned to each rank component of the LoRA experts for more nuanced discipline-specific patterns. 

Figure~\ref{fig:analysis2}(c) shows that the shared-expert design yields only marginal gains, whereas sharing the A matrix leads to substantial improvements. Expert activation visualizations in Figure~\ref{fig:expert} explain this gap. While shared experts in standard MoE~\citep{liu2024deepseekv3} are meant to capture common knowledge, in LoRA-MoE they dilute specialization, as indicated by near-uniform activations across disciplines (except chemistry).
In contrast, sharing the A matrix aligns LoRA experts in a common latent space, enabling clearer specialization (e.g., expert 2 for mathematics and expert 4 for chemistry). Additionally, rank-wise routing, despite finer-grained control, neither improves performance nor remains stable in data-scarce disciplines (Figure~\ref{fig:analysis2}(c); Appendix \ref{sec:add_res}), likely due to excessive routing flexibility destabilizing expert selection.
Overall, these findings highlight a share-then-specialize law, where early parameter sharing provides a common representation foundation that subsequently promotes clearer expert specialization and improved multi-disciplinary performance.

\section{Final Multi-Discipline Learning Recipe}

Based on our previous analysis, we formulate the following multi-discipline fine-tuning recipe:
\begin{itemize}
    \item \textbf{Data Scarcity Mitigation:} To address performance degradation caused by data-scarce disciplines, we upsample the biology and geography training corpora to 100K samples each, following the diversity-aware upsampling strategy shown to be most effective.
    \item \textbf{Enhancing Instruction Following:} Given its substantial contribution to multi-disciplinary performance, we incorporate the general-domain instruction-following enhancement data from InternLM \citep{cai2024internlm2} into the multi-discipline training corpus.
    \item \textbf{Fine-Tuning Strategy:} We adopt full-model tuning and LoRA-MoE as the final fine-tuning approaches. For LoRA-MoE, we apply the shared LoRA A matrix design with token-level and layer-wise routing.
\end{itemize}
All models are fine-tuned using the setup in Section~\ref{sec:setup}, with results summarized in Table~\ref{tab:recipe}. In addition to discipline-wise accuracy, we report the delta performance $\Delta m$~\citep{agiza2024cv}, measuring the average per-task performance drop relative to the single-discipline LoRA baseline, along with the percentage of trainable parameters.
As shown in Table~\ref{tab:recipe}, both full fine-tuning and LoRA-MoE trained with the proposed recipe achieve substantial improvements over the original setting, validating the effectiveness of the combined strategies. Notably, LoRA-MoE largely closes the gap to full fine-tuning, attaining comparable in-domain accuracy while using only a small fraction of trainable parameters. Moreover, LoRA-MoE exhibits stronger out-of-domain generalization on MMLU \citep{hendrycks2020mmlu}, achieving these gains with only ~15\% of the GPU hours required by full fine-tuning.

\section{Conclusion}
We present the first systematic analysis of multi-disciplinary LLM fine-tuning, revealing significantly higher training variability than single-discipline settings and distilling four empirical laws which are Balance-then-Diversity, Merge-then-Align, Optimize-then-Scale, and Share-then-Specialize, that govern effective cross-domain learning. Together, these principles provide a practical recipe for building robust and generalizable multi-discipline scientific LLMs. 

\clearpage
\section*{Limitations}

While our study provides comprehensive insights into multi-discipline fine-tuning of LLMs, several limitations should be acknowledged. Firstly, our multi-discipline corpus is constructed by aggregating existing single-discipline datasets. However, these datasets may not fully capture the diversity or specific characteristics of their respective disciplines, particularly in fields with complex or nuanced knowledge structures. As such, the findings and observed learning patterns may vary when applied to other disciplines or to datasets with different properties. Future work should explore broader discipline coverage and incorporate more diverse datasets to enhance representativeness and generalizability. Due to the high computational cost associated with full-model fine-tuning, our analysis is limited to key scenarios, specifically focusing on the multi-discipline setting and its optimized configuration. We do not provide scaling analyses for full fine-tuning across all data volumes, which may limit our understanding of its behavior in lower-resource or discipline-specific settings. Further investigations using more efficient full fine-tuning approximations or resource-scalable evaluations are needed. Lastly, LLM fine-tuning is known to exhibit performance variability due to optimization instability and data randomness. Although we conducted multiple runs for cases showing suspicious outliers to mitigate this concern, inherent variability remains a potential limitation. More robust methods, such as uncertainty estimation or larger-scale repeated trials, could be explored in future studies to further strengthen the reliability of the findings.

\bibliography{custom}

\clearpage
\appendix
\section{Additional Experiment Setup}
\label{sec:add}

\begin{table*}[thb]
  \centering
    \begin{tabular}{p{4cm}<{\raggedright\arraybackslash}|c|cccccc}
      \toprule

      \multicolumn{1}{l}{\textbf{Method}}
      & \multicolumn{1}{c}{\textbf{Data}}
      & \multicolumn{1}{c}{\textbf{Chem}}
      & \multicolumn{1}{c}{\textbf{Med}}
      & \multicolumn{1}{c}{\textbf{Bio}}
      & \multicolumn{1}{c}{\textbf{Math}}
      & \multicolumn{1}{c}{\textbf{Geo}}
      & \multicolumn{1}{c}{\textbf{Avg}} \\

      \midrule
      Discipline LoRA
        & \multirow{4}{*}{10 \%}
        & 0.495 & \textbf{0.576} & 0.656 & 0.836 & 0.746 & 0.662 \\
      LoRA
        & 
        & 0.541 & 0.491 & 0.669 & \textbf{0.854} & 0.724 & 0.656 \\
      LoRA-MoE
        & 
        & \textbf{0.571} & 0.471 & 0.489 & \textbf{0.854} & 0.703 & 0.618 \\
      LoRA-Comp
        & 
        & 0.537 & 0.570 & \textbf{0.691} & 0.847 & \textbf{0.750} & \textbf{0.679} \\

      \hline
      Discipline LoRA
        & \multirow{4}{*}{20 \%}
        & 0.490 & \textbf{0.579} & 0.640 & 0.840 & 0.748 & 0.659 \\
      LoRA
        & 
        & 0.552 & 0.544 & \textbf{0.713} & 0.853 & 0.732 & 0.679 \\
      LoRA-MoE
        & 
        & \textbf{0.560} & 0.510 & 0.599 & 0.847 & 0.734 & 0.650 \\
      LoRA-Comp
        & 
        & 0.538 & 0.576 & 0.692 & \textbf{0.862} & \textbf{0.755} & \textbf{0.685} \\

      \hline
      Discipline LoRA
        & \multirow{4}{*}{50 \%}
        & 0.495 & 0.559 & 0.694 & 0.841 & 0.747 & 0.667 \\
      LoRA
        & 
        & 0.539 & \textbf{0.580} & \textbf{0.700} & 0.855 & \textbf{0.755} & \textbf{0.686} \\
      LoRA-MoE
        & 
        & \textbf{0.556} & 0.574 & 0.556 & 0.840 & 0.735 & 0.652 \\
      LoRA-Comp
        & 
        & 0.528 & 0.553 & 0.660 & \textbf{0.857} & 0.740 & 0.668 \\ 

      \hline
      Discipline LoRA
        & \multirow{4}{*}{70 \%}
        & 0.533 & \textbf{0.563} & 0.694 & 0.846 & 0.743 & 0.676 \\
      LoRA
        & 
        & 0.557 & \textbf{0.563} & 0.642 & 0.862 & 0.667 & 0.658 \\
      LoRA-MoE
        & 
        & \textbf{0.576} & 0.514 & 0.585 & 0.859 & 0.733 & 0.653 \\
      LoRA-Comp
        & 
        & 0.549 & 0.548 & \textbf{0.701} & \textbf{0.864} & \textbf{0.747} & \textbf{0.682} \\ 
        
      \hline
      Discipline LoRA
        & \multirow{5}{*}{100 \%}
        & 0.530 & \textbf{0.581} & 0.713 & 0.845 & 0.745 & 0.683 \\
      FT
        & 
        & \textbf{0.619} & 0.545 & 0.646 & 0.830 & 0.707 & 0.669 \\
      LoRA
        & 
        & 0.553 & 0.549 & 0.702 & 0.836 & 0.745 & 0.677 \\
      LoRA-MoE
        & 
        & 0.535 & 0.550 & 0.683 & \textbf{0.847} & 0.741 & 0.671 \\
      LoRA-Comp
        & 
        & 0.564 & 0.555 & \textbf{0.720} & 0.846 & \textbf{0.750} & \textbf{0.687} \\
      \bottomrule
    \end{tabular}
  \caption{Full evaluation results for all fine-tuning methods under all tested data scales. For each data scale, we highlight the \textbf{best value}.}
  \label{tab:full_result}
  \vspace{-5pt} 
\end{table*}

\begin{figure*}[thb]
     \centering
     \includegraphics[width=0.95\linewidth]{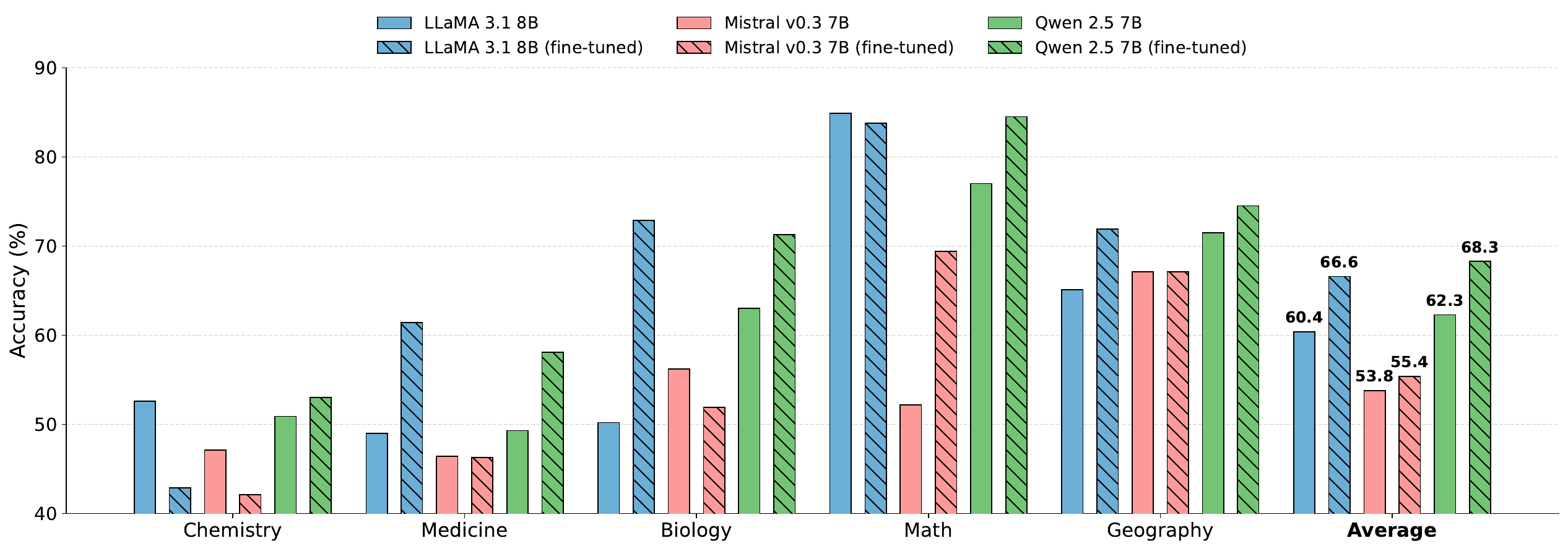}
     \caption{Base models' performance before and after fine-tuning for each discipline by preliminary scientific LoRA. }
     \label{fig:base}
\end{figure*}

\begin{table*}[thb]
  \centering
    \begin{tabular}{p{4cm}<{\raggedright\arraybackslash}|c|cccccc}
      \toprule

      \multicolumn{1}{l}{\textbf{Method}}
      & \multicolumn{1}{c}{\textbf{Upsample}}
      & \multicolumn{1}{c}{\textbf{Chem}}
      & \multicolumn{1}{c}{\textbf{Med}}
      & \multicolumn{1}{c}{\textbf{Bio}}
      & \multicolumn{1}{c}{\textbf{Math}}
      & \multicolumn{1}{c}{\textbf{Geo}}
      & \multicolumn{1}{c}{\textbf{Avg}} \\

      \midrule

      LoRA
        & \multirow{3}{*}{NA}
        & 0.553 & 0.549 & 0.702 & 0.836 & 0.745 & 0.677 \\
      LoRA-MoE
        & 
        & 0.535 & 0.550 & 0.683 & \textbf{0.847} & 0.741 & 0.671 \\
      LoRA-Comp
        & 
        & \textbf{0.564} & \textbf{0.555} & \textbf{0.720} & 0.846 & \textbf{0.750} & \textbf{0.687} \\

      \hline
      
      LoRA (diverse)
        & \multirow{6}{*}{100K}
        & \textbf{0.570} & \textbf{0.586} & \textbf{0.765} & 0.842 & 0.720 & \textbf{0.697} \\
      LoRA (unique)
        & 
        & 0.556 & 0.580 & 0.588 & \textbf{0.843} & \textbf{0.749} & 0.663 \\

      \cline{1-1}\cline{3-8}
    
      LoRA-MoE (diverse)
        & 
        & \textbf{0.564} & \textbf{0.577} & \textbf{0.771} & 0.844 & \textbf{0.725} & \textbf{0.696} \\
      LoRA-MoE (unique) 
        & 
        & 0.559 & 0.497 & 0.598 & \textbf{0.853} & 0.712 & 0.644 \\ 

      \cline{1-1}\cline{3-8}
      
      LoRA-Comp (diverse)
        & 
        & 0.561 & 0.560 & \textbf{0.715} & 0.845 & 0.753 & 0.687 \\
      LoRA-Comp (unique) 
        & 
        & \textbf{0.562} & \textbf{0.566} & \textbf{0.715} & \textbf{0.849} & \textbf{0.758} & \textbf{0.690} \\ 
        
      \hline
      
      LoRA (diverse)
        & \multirow{6}{*}{300K}
        & \textbf{0.541} & \textbf{0.547} & \textbf{0.688} & 0.829 & \textbf{0.728} & \textbf{0.667} \\
      LoRA (unique)
        & 
        & 0.511 & 0.522 & 0.511 & \textbf{0.839} & 0.515 & 0.580 \\

      \cline{1-1}\cline{3-8}
    
      LoRA-MoE (diverse)
        & 
        & \textbf{0.536} & 0.535 & 0.653 & \textbf{0.845} & 0.734 & 0.661 \\
      LoRA-MoE (unique) 
        & 
        & 0.535 & \textbf{0.556} & \textbf{0.682} & 0.838 & \textbf{0.739} & \textbf{0.670} \\ 
        
      \cline{1-1}\cline{3-8}
      
      LoRA-Comp (diverse)
        & 
        & 0.566 & 0.552 & \textbf{0.730} & \textbf{0.848} & \textbf{0.756} & 0.690 \\
      LoRA-Comp (unique) 
        & 
        & \textbf{0.567} & \textbf{0.556} & 0.727 & 0.846 & 0.751 & \textbf{0.691} \\ 
        
      \bottomrule
    \end{tabular} 
  \caption{Full evaluation results for different upsampling strategy. For each upsampling setting and fine-tuning method, we highlight the \textbf{best value}.}
  \label{tab:upsample_result}
  \vspace{-5pt}
\end{table*}

\textbf{Multi-Discipline Setting} 
To expand the medicine discipline dataset and align it with the scale and recency of other disciplines in our multi-discipline setup, we aggregate data from a range of existing sources \citep{vilares2019head, han2023medalpaca, pal2022medmcqa, jin2021disease, ben2019question, naturalinstructions, OpenGPT, jin2019pubmedqa, li2023chatdoctor, BenAbacha:MEDINFO19, bodenreider2004unified, zhang2023alpacare}. The resulting dataset includes:

\begin{itemize}
    \item 2,657 multiple-choice questions from HeadQA \citep{vilares2019head}
    \item 120,765 multiple-choice questions from MedMCQA \citep{pal2022medmcqa}
    \item 10,178 multiple-choice questions from MedQA \citep{jin2021disease}
    \item 119,486 real doctor-patient conversations from ChatDoctor \citep{li2023chatdoctor}
    \item 16,407 QA pairs from MedQuAD \citep{ben2019question}
    \item 500 biomedical questions from PubMedQA \citep{jin2019pubmedqa}
    \item 689 consumer drug-related questions from MedicationQA \citep{BenAbacha:MEDINFO19}
    \item 79,245 medical QA pairs derived from UMLS \citep{bodenreider2004unified}
    \item 24,665 QA pairs from OpenGPT \citep{OpenGPT}
    \item 52,002 machine-generated QA pairs from AlpacaRE \citep{zhang2023alpacare}
    \item A subset of Medical Meadow dataset\citep{han2023medalpaca}, including health advice, flashcards, PubMed-based causal data
    \item Medical-related tasks from the Natural Instructions v2.8 dataset \citep{naturalinstructions}, including tasks 179, 181, 1369, 1447, 1449, 1485, 1487, 1495, and 1645
\end{itemize}

The final aggregation comprises 490,766 samples, ensuring sufficient data diversity and scale to support robust multi-discipline fine-tuning in the medical domain.

\textbf{Evaluation Setting}
We evaluate models using the lm-evaluation-harness framework \cite{eval-harness} with two modes. For math, which involves direct-answer questions, the model generates output until the end-of-sequence token, and the final answer is extracted and compared to the reference. For the remaining disciplines, which use multiple-choice formats, the model selects the option with the highest log-likelihood \cite{hendrycks2020mmlu}, which is then compared to the ground truth. 

\textbf{Finetuning Setting}
We provide additional details of our fine-tuning setup. For full-model tuning, all parameters are updated using a learning rate of $7 \times 10^{-6}$, weight decay of $0.1$, and a warm-up ratio of $0.05$, trained for 1 epoch. For PEFT methods, we use a learning rate of $1 \times 10^{-4}$, weight decay of $0.01$, and a warm-up ratio of $0.1$, also trained for 1 epoch. Each LoRA module is configured with rank 16 and a scaling factor of 32, applied to all attention and feedforward layers. The gating network in LoRA-MoE and LoRA-Comp is implemented as a single-layer MLP that predicts layer-wise gating weights based on the layer input. Both LoRA-MoE and LoRA-Comp use five experts, aiming for one per discipline, with each expert configured identically to the single LoRA setup. All models are trained using the AdamW optimizer \citep{loshchilov2017adamw} with a linear learning rate scheduler, an effective batch size of 128, and are implemented using HuggingFace Transformers \citep{wolf2020huggingface} and DeepSpeed \citep{rasley2020deepspeed} on 32 NVIDIA A800 GPUs.

\section{Additional Results}
\label{sec:add_res}

In addition to the visualizations in the main manuscript, we present comprehensive evaluation results in the following tables: Table~\ref{tab:full_result} (fine-tuning methods across all data scales), Table~\ref{tab:upsample_result} (upsampling strategies), Table~\ref{tab:instruct_result} (instruction-following enhancement), Table~\ref{tab:scale_result} (trainable parameter scaling), and Table~\ref{tab:model_result} (model architectual design).

\section{Base Model Selection}
\label{sec:base}

In Figure \ref{fig:base}, we present the effects of preliminary scientific LoRA fine-tuning across different base models(LLaMA 3.1 7B, Mistral v0.3 7B, Qwen 2.5 7B) and scientific disciplines(chemistry, medicine, biology, math, geography). The fine-tuned models achieve higher average performance than their counterparts. Performance differs across base models, with Qwen 2.5 7B exhibiting higher overall performance and improvements across disciplines.

\begin{table*}[th]
  \centering
    \begin{tabular}{p{4.5cm}<{\raggedright\arraybackslash}|c|cccccc}
      \toprule

      \multicolumn{1}{l}{\textbf{Method}}
      & \multicolumn{1}{c}{\textbf{Data}}
      & \multicolumn{1}{c}{\textbf{Chem}}
      & \multicolumn{1}{c}{\textbf{Med}}
      & \multicolumn{1}{c}{\textbf{Bio}}
      & \multicolumn{1}{c}{\textbf{Math}}
      & \multicolumn{1}{c}{\textbf{Geo}}
      & \multicolumn{1}{c}{\textbf{Avg}} \\

      \midrule
      
      LoRA 
        & \multirow{6}{*}{70\%}
        & 0.557 & 0.563 & 0.642 & \textbf{0.862} & 0.667 & 0.658 \\
      LoRA (mix gen)
        & 
        & \textbf{0.558} & \textbf{0.597} & \textbf{0.766} & 0.844 & \textbf{0.761} & \textbf{0.705} \\

      \cline{1-1}\cline{3-8}
        
      LoRA-MoE
        & 
        & \textbf{0.576} & 0.514 & 0.585 & \textbf{0.859} & 0.733 & 0.653 \\
      LoRA-MoE (mix gen) 
        & 
        & 0.572 & \textbf{0.594} & \textbf{0.790} & 0.854 & \textbf{0.753} & \textbf{0.713} \\

      \cline{1-1}\cline{3-8}
      
      LoRA-Comp
        & 
        & 0.549 & 0.548 & 0.701 & 0.864 & 0.747 & 0.682 \\ 
      LoRA-Comp (mix gen) 
        & 
        & \textbf{0.551} & \textbf{0.554} & \textbf{0.720} & \textbf{0.870} & \textbf{0.753} & \textbf{0.690} \\ 
        
      \hline

      LoRA
        & \multirow{6}{*}{100\%}
        & 0.553 & 0.549 & 0.702 & 0.836 & 0.745 & 0.677 \\
      LoRA (mix gen)
        & 
        & \textbf{0.571} & \textbf{0.600} & \textbf{0.781} & \textbf{0.851} & \textbf{0.758} & \textbf{0.712} \\

      \cline{1-1}\cline{3-8}
        
      LoRA-MoE
        & 
        & 0.535 & 0.550 & 0.683 & \textbf{0.847} & 0.741 & 0.671 \\
      LoRA-MoE (mix gen)
        & 
        & \textbf{0.542} & \textbf{0.600} & \textbf{0.805} & 0.839 & \textbf{0.754} & \textbf{0.708} \\

      \cline{1-1}\cline{3-8}
        
      LoRA-Comp
        & 
        & 0.564 & 0.555 & 0.720 & \textbf{0.846} & \textbf{0.750} & 0.687 \\
      LoRA-Comp (mix gen) 
        & 
        & \textbf{0.567} & \textbf{0.569} & \textbf{0.730} & 0.845 & \textbf{0.750} & \textbf{0.692} \\
      \bottomrule
    \end{tabular}
  \caption{Full evaluation results for improving instruction following ability. For each data scale and fine-tuning method, we highlight the \textbf{best value}.}
  \label{tab:instruct_result}
  \vspace{-5pt} 
\end{table*}

\begin{table*}[thb]
  \centering
    \begin{tabular}{p{4.5cm}<{\raggedright\arraybackslash}|c|cccccc}
      \toprule

      \multicolumn{1}{l}{\textbf{Method}}
      & \multicolumn{1}{c}{\textbf{Data}}
      & \multicolumn{1}{c}{\textbf{Chem}}
      & \multicolumn{1}{c}{\textbf{Med}}
      & \multicolumn{1}{c}{\textbf{Bio}}
      & \multicolumn{1}{c}{\textbf{Math}}
      & \multicolumn{1}{c}{\textbf{Geo}}
      & \multicolumn{1}{c}{\textbf{Avg}} \\

      \midrule
      
      LoRA (r=16)
        & \multirow{6}{*}{70\%}
        & 0.557 & 0.563 & 0.642 & \textbf{0.862} & 0.667 & 0.658 \\
      LoRA (r=80)
        & 
        & 0.556 & \textbf{0.588} & 0.671 & 0.847 & 0.743 & 0.681 \\
      LoRA (r=160)
        & 
        & \textbf{0.585} & 0.538 & \textbf{0.755} & 0.848 & \textbf{0.748} & \textbf{0.695} \\

      \cline{1-1}\cline{3-8}
        
      LoRA-MoE (e=5)
        & 
        & 0.576 & 0.514 & \textbf{0.585} & \textbf{0.859} & 0.733 & \textbf{0.653} \\
      LoRA-MoE (e=10)
        & 
        & \textbf{0.594} & \textbf{0.566} & 0.455 & 0.841 & \textbf{0.757} & 0.643 \\
      LoRA-MoE (e=20)
        & 
        & 0.591 & 0.537 & 0.550 & 0.846 & 0.731 & 0.651 \\

      \hline
      LoRA (r=16)
        & \multirow{6}{*}{100\%}
        & 0.553 & 0.549 & 0.702 & 0.836 & \textbf{0.745} & 0.677 \\
      LoRA (r=80)
        & 
        & \textbf{0.562} & 0.542 & 0.547 & 0.830 & 0.734 & 0.643 \\
      LoRA (r=160)
        & 
        & 0.555 & \textbf{0.598} & \textbf{0.749} & \textbf{0.844} & \textbf{0.745} & \textbf{0.698} \\

      \cline{1-1}\cline{3-8}
        
      LoRA-MoE (e=5)
        & 
        & 0.535 & 0.550 & 0.683 & \textbf{0.847} & 0.741 & 0.671 \\
      LoRA-MoE (e=10)
        & 
        & \textbf{0.561} & \textbf{0.554} & \textbf{0.727} & 0.821 & 0.751 & \textbf{0.683} \\
      LoRA-MoE (e=20)
        & 
        & 0.557 & 0.552 & 0.644 & 0.828 & \textbf{0.757} & 0.668 \\
      \bottomrule
    \end{tabular} 
  \caption{Full evaluation results for scaling model trainable parameters. For each data scale and fine-tuning method, we highlight the \textbf{best value}.}
  \label{tab:scale_result}
  \vspace{-15em}
\end{table*}

\begin{table*}[thb]
  \centering
    \begin{tabular}{p{4.5cm}<{\raggedright\arraybackslash}|c|cccccc}
      \toprule

      \multicolumn{1}{l}{\textbf{Method}}
      & \multicolumn{1}{c}{\textbf{Data}}
      & \multicolumn{1}{c}{\textbf{Chem}}
      & \multicolumn{1}{c}{\textbf{Med}}
      & \multicolumn{1}{c}{\textbf{Bio}}
      & \multicolumn{1}{c}{\textbf{Math}}
      & \multicolumn{1}{c}{\textbf{Geo}}
      & \multicolumn{1}{c}{\textbf{Avg}} \\

      \midrule
      
      LoRA-MoE
        & \multirow{4}{*}{70 \%}
        & 0.576 & \textbf{0.514} & 0.585 & 0.859 & 0.733 & 0.653 \\
      LoRA-MoE (shared expert)
        & 
        & 0.531 & 0.502 & \textbf{0.699} & \textbf{0.861} & 0.741 & \textbf{0.667} \\
      LoRA-MoE (shared A)
        & 
        & 0.578 & 0.476 & 0.672 & 0.848 & \textbf{0.746} & 0.664 \\
      LoRA-MoE (rank route)
        & 
        & \textbf{0.588} & 0.480 & 0.431 & 0.848 & 0.734 & 0.616 \\
        
      \hline
      LoRA-MoE
        & \multirow{4}{*}{100 \%}
        & 0.535 & 0.550 & 0.683 & 0.847 & 0.741 & 0.671 \\
      LoRA-MoE (shared expert)
        & 
        & 0.568 & 0.532 & 0.677 & 0.850 & 0.751 & 0.676 \\
      LoRA-MoE (shared A)
        & 
        & 0.567 & \textbf{0.594} & \textbf{0.772} & \textbf{0.852} & \textbf{0.753} & \textbf{0.708} \\
      LoRA-MoE (rank route)
        & 
        & \textbf{0.575} & 0.496 & 0.422 & 0.813 & 0.620 & 0.585 \\
      \bottomrule
    \end{tabular} 
  \caption{Full evaluation results for different architecture design. For each data scale, we highlight the \textbf{best value}.}
  \label{tab:model_result}
  \vspace{-5pt}
\end{table*}

\end{document}